\documentclass[10pt,twocolumn,letterpaper]{article}

\usepackage{cvpr}              
\usepackage{paralist}

\usepackage{booktabs}
\usepackage{multirow}
\usepackage[normalem]{ulem}
\useunder{\uline}{\ul}{}
\usepackage{threeparttable}
\usepackage{array}
\usepackage{graphicx}
\usepackage{subcaption}
\usepackage{adjustbox}
\usepackage{stfloats}
\usepackage{tabularx}
\usepackage{xcolor}
\usepackage{colortbl}
\usepackage[accsupp]{axessibility}  

\newcommand*\rot{\rotatebox{70}}

\usepackage{pifont}
\newcommand{\cmark}{\ding{51}}%
\newcommand{\xmark}{\ding{55}}%

\usepackage{arydshln}

\definecolor{cvprblue}{rgb}{0.21,0.49,0.74}
\usepackage[pagebackref,breaklinks,colorlinks,allcolors=cvprblue]{hyperref}

\def\paperID{***} 
\def\confName{CVPR}
\def\confYear{2026}

\title{RHO: Robust Holistic OSM-Based Metric Cross-View Geo-Localization}

\author{Junwei Zheng$^{1,3,*,\dag}$ \quad 
Ruize Dai$^{1,*}$ \quad 
Ruiping Liu$^{1}$ \quad 
Zichao Zeng$^{1,4}$ \quad 
Yufan Chen$^{1}$ \quad  \\
Fangjinhua Wang$^{3}$ \quad
Kunyu Peng$^{1,5}$ \quad 
Kailun Yang$^{2}$ \quad 
Jiaming Zhang$^{2,\dag}$ \quad 
Rainer Stiefelhagen$^{1}$\\
\normalsize
$^1$Karlsruhe Institute of Technology
\normalsize \quad
$^2$Hunan University
\normalsize \quad
$^3$ETH Zurich  \\
\normalsize \quad
$^4$UCL
\normalsize \quad
$^5$INSAIT, Sofia University ``St. Kliment Ohridski''
}

\begin{document}
\maketitle
{
  \renewcommand{\thefootnote}
    {\fnsymbol{footnote}}
  \footnotetext[1]{Equal contribution.}
  \footnotetext[2]{Correspondence: \href{mailto:junwei.zheng@kit.edu}{\textcolor{blue}{junwei.zheng@kit.edu}}, \href{mailto:jiamingzhang@hnu.edu.cn}{\textcolor{blue}{jiamingzhang@hnu.edu.cn}}.}
}

\begin{abstract}
\label{sec:abstract}
Metric Cross-View Geo-Localization (MCVGL) aims to estimate the $3$-DoF camera pose (position and heading) by matching ground and satellite images. In this work, instead of pinhole and satellite images, we study robust MCVGL using holistic panoramas and OpenStreetMap (OSM). To this end, we establish a large-scale MCVGL benchmark dataset, \textbf{CV-RHO}, with over $2.7$M images under different weather and lighting conditions, as well as sensor noise. Furthermore, we propose a model termed \textbf{RHO} with a two-branch \textbf{Pin-Pan} architecture for accurate visual localization. A Split-Undistort-Merge (\textbf{SUM}) module is introduced to address the panoramic distortion, and a Position-Orientation Fusion (\textbf{POF}) mechanism is designed to enhance the localization accuracy. Extensive experiments prove the value of our CV-RHO dataset and the effectiveness of the RHO model, with a significant performance gain up to $20\%$ compared with the state-of-the-art baselines.
Project page: \url{https://github.com/InSAI-Lab/RHO}.
\end{abstract}    
\section{Introduction}
\label{sec:intro}

Cross-View Geo-Localization (CVGL) is one of the fundamental tasks in the computer vision domain.
It can be further refined into two sub-directions: Large-Area Cross-View Geo-Localization (LCVGL)~\cite{zhu2022transgeo,deuser2023sample4geo,fervers2024statewide,ye2025where} and Metric Cross-View Geo-Localization (MCVGL)~\cite{xia2025loc2,tong2025geodistill,tian2025generalizing,xia2024adapting}.
LCVGL starts from a large (\eg, city-scale) search region and finds a rough estimate of the camera position.
It typically uses an image retrieval approach and therefore does not predict orientation or reach high metric accuracy.
MCVGL, on the other hand, starts from a noisy GPS prior (\eg, $\pm$50m error~\cite{karaim2018gnss}) and determines the position and orientation with higher accuracy at meter level by matching ground and satellite images, which is more challenging and valuable in real-world applications, \eg, autonomous driving~\cite{xia2025fg2,fervers2023uncertainty} and remote sensing~\cite{yao2024uav,zhao2024transfg}. 

\begin{figure}[!t]
    \centering
    \includegraphics[width=\linewidth]{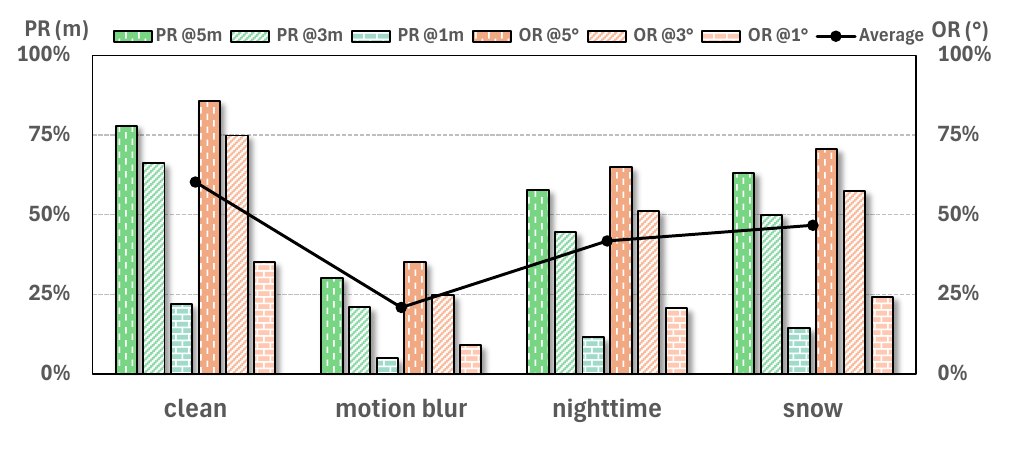}
    \vskip -1.3em
    \caption{OrienterNet~\cite{sarlin2023orienternet} fails under adverse conditions.
    PR and OR stand for Position Recall and Orientation Recall.}
    \label{fig:performance-degradation}
    \vskip -4ex
\end{figure}

In this work, we study robust MCVGL using panoramas and OpenStreetMap (OSM).
The motivation is threefold.
(1) First, the majority of existing studies in the community focus on addressing the MCVGL problem under ideal conditions (\ie, sunny with sufficient light).
This assumption doesn't always hold.
The effects of varying weather and lighting conditions, as well as sensor noise, on MCVGL remain largely underexplored.
Fig.~\ref{fig:performance-degradation} showcases that models trained on the dataset under ideal conditions fail at predicting camera poses under noisy conditions with a dramatic performance degradation.
(2) Second, in contrast to pinhole images, panoramic images offer richer visual information~\cite{zheng2024ops,zheng2025spr,hu2024dmamba,fan2026plm,liu2026if}, which facilitates more accurate estimation of the camera’s position and orientation.
(3) Apart from enhancing robustness and utilizing holistic panoramic representations, incorporating OSM also plays an important role in addressing the MCVGL task.
Since satellite imagery is updated infrequently, MCVGL fails when attempting to match newly captured ground images with outdated satellite images.
Another advantage of utilizing OSM over satellite imagery lies in its significantly lower storage requirements.
While satellite imagery occupies approximately $75$MB of storage per $1$km$^{2}$, OSM data requires only about one-fifteenth of the space with only $4.8$MB~\cite{sarlin2023orienternet}.

To evaluate the robustness of MCVGL, we create \textbf{CV-RHO} dataset, with over $2.7$M images under varying conditions and perturbations, including \emph{rain}, \emph{snow}, \emph{fog}, and \emph{nighttime} scenarios, as well as variations such as \emph{overexposure}, \emph{underexposure}, and \emph{motion blur}, as illustrated in Fig.~\ref{fig:data_samples}.
It's worth noting that all generated images are structurally consistent with the original images.
CV-RHO is the first large-scale benchmark dataset aiming at studying the robustness of OSM-based MCVGL, providing a promising step towards the robustness and generalization of the CVGL task.
Furthermore, we propose a model termed \textbf{RHO} with a two-branch \textbf{Pin-Pan} architecture.
The panoramic branch takes the 360{\textdegree}-FoV images as input and predicts the probability distribution of the camera positions.
To handle the panoramic distortion due to the equirectangular projection, a Split-Undistort-Merge (\textbf{SUM}) module is introduced and integrated into the panoramic branch.
SUM firstly splits one panorama into three 120{\textdegree}-FoV pinhole images and then undistorts the panoramic deformation.
All three pinholes go through the image encoder and the Bird's-Eye-View (BEV) projection, resulting in three BEV neural maps, which are finally merged in the BEV space.
The fused BEV neural map is utilized to align with the OSM data in order to estimate the distribution of camera positions.
This matching results in a three-dimensional (\ie, H, W, Angles) distribution volume. 
The pinhole branch, on the other hand, estimates the camera heading using the pinhole image derived from the panorama.
A Position-Orientation Fusion (\textbf{POF}) mechanism is designed to fuse the panoramic and pinhole distribution volumes.
The panoramic volume is aggregated along its orientational dimension to refine the pinhole volume, while the pinhole volume is aggregated along its spatial dimensions to enrich the panoramic volume.
Extensive experiments on the CV-RHO dataset verify the effectiveness of our RHO model with a significant boost of up to 20\% under different conditions, outperforming other state-of-the-art baselines. To summarize, our contributions are: 
\begin{compactitem}
    \item We create the first large-scale OSM-based MCVGL dataset with over $2.7$M images, \textbf{CV-RHO}, to study the robustness of MCVGL under different weather and lighting conditions, as well as sensor noise, including rain, snow, fog, night, overexposure, underexposure, and motion blur.
    \item We propose a novel model termed \textbf{RHO} with a two-branch \textbf{Pin-Pan} architecture. The panoramic branch aims to predict the camera position, while the pinhole branch is for the orientation estimation.
    \item We introduce a \textbf{Split-Undistort-Merge} module to handle the panoramic distortion due to the equirectangular projection.
    \item We design a \textbf{Position-Orientation Fusion} mechanism to mutually enhance the probability distribution of camera position and heading.
    \item Extensive experiments on the CV-RHO dataset prove the value of the CV-RHO and the effectiveness of the RHO model with a significant performance gain up to 20\% over the baseline.
\end{compactitem}

\section{Related Work}
\label{sec:related_work}

\noindent\textbf{OSM-Based Cross-View Geo-Localization}.
Satellite-based CVGL~\cite{wang2024vfa,xia2025fg2,fervers2023uncertainty,xia2023ccvpe,klammer2024bevloc,li2024unleashing,zeng2022geo,wang2023fine,hu2022beyond,vyas2022gama} refers to matching the ground and satellite images to predict the position and orientation of the current viewpoint.
In contrast to satellite-based CVGL, OSM-based CVGL leverages the OpenStreetMap (OSM) instead of the satellite images during the matching process.
OrienterNet~\cite{sarlin2023orienternet}, as the seminal work, first introduces the concept of OSM-based CVGL.
OMSLoc~\cite{liao2024osmloc} proposes a geometry-guided BEV transform with the foundation model’s depth prior, and an auxiliary semantic alignment task that enhances the scene understanding ability of the model.
Similar to OSMLoc, DP-Loc~\cite{kim2024dploc} leverages the prior depth for boosting the localization accuracy.
MapLocNet~\cite{wu2024maplocnet} applies a coarse-to-fine feature registration mechanism to reduce the inference latency.
U-BEV~\cite{camiletto2024ubev} presents a lightweight architecture that is geometrically constrained and leverages the height of the points from the ground instead of their depth from the camera.
SegLocNet~\cite{zhou2025seglocnet} proposes a multimodal localization network that achieves highly accurate localization in GPS-denied areas by incorporating the BEV segmentation and 2D navigation maps.
DiffVL~\cite{gao2025diffvl} integrates diffusion models to denoise noisy GPS trajectories for OSM-based CVGL, achieving better localization accuracy.
In this work, we delve into the OSM-based CVGL since OSM is updated more frequently than satellite imagery and is storage-friendly.

\noindent\textbf{Robust Cross-View Geo-Localization}.
Most of the existing works~\cite{lu2025gleam,ding2025cross_3dgs,wang2025bevsplat,li2025vageo,shi2022beyond,shi2020optimal,yang2021cross,zhang2023cross,wang2022satellite,wang2023view} assume that the cross-view matching is under ideal conditions, which doesn't always hold in the real world.
Robust CVGL is underexplored.
SSPT~\cite{fan2024sspt} develops a multi-modal data augmentation strategy for Unmanned Aerial Vehicle (UAV) image datasets, broadening the diversity and complexity of the training data.
Weln~\cite{lv2024weln} enhances the model robustness by expanding the University-1652~\cite{zheng2020university} dataset with nine different weather conditions added. 
A benchmark dataset~\cite{zhang2024benchmarking} for evaluating the robustness of CVGL models is proposed by introducing 16 common types of data corruption.
MCGF~\cite{feng2024mcgf} introduces a multi-weather CVGL framework designed to dynamically adapt to unseen weather conditions.
ConGeo~\cite{mi2024congeo} enhances robustness and consistency in feature representations to improve a model's invariance to orientation and its resilience to FoV variations.
CVD~\cite{li2025robust_disentanglement} explicitly disentangles content and viewpoint factors for robust CVGL.
\cite{guan2025dual_branch} presents a dual-branch transformer framework with gradient-aware weighting feature alignment for robust CVGL.
In addition to a multi-dimensional feature matching mechanism, \cite{wang2025towards_robust} introduces a center-focused attention mechanism to improve the matching accuracy and robustness.
However, there are no benchmarks and methods targeting the OSM-based MCVGL, an emerging and highly promising frontier for visual localization.
To bridge this gap, we propose the first large-scale OSM-based MCVGL dataset with over $2.7$M images, CV-RHO, as well as training a robust model, RHO, against diverse perturbations.

\noindent\textbf{Panoramic Cross-View Geo-Localization}.
Narrow-FoV ground images~\cite{rodrigues2023semgeo,shi2023boosting,shugaev2024arcgeo,xie2025self,pillai2024garet,shore2024bev} provide limited visual information to generate BEV features on the CVGL task, resulting in low accuracy in visual localization.
Panoramas~\cite{zhu2021revisiting,yuan2024cross,xia2025auxgeo,hu2025oriloc}, on the other hand, enable a holistic understanding of the environments and generate $360^\circ$-FoV BEV features, achieving more accurate cross-view matching.
SIRNet~\cite{lu2022s} decomposes the complex learning process into several refinement steps while adapting the refinement steps specifically for each panoramic input.
GeoDTR~\cite{zhang2023geodtr} explicitly disentangles geometric information from raw features and learns the spatial correlations among visual features from aerial and panoramic ground pairs.
GeoDTR+~\cite{zhang2024geodtr+}, built upon GeoDTR, models the spatial configuration of both panoramic and aerial images through the geometric layout extraction mechanism.
TriViTs~\cite{ahn2024bridging} is designed to leverage both spatially aligned and original cross-view images.
DSTG~\cite{liang2025dstg} enhances panoramic image representation through a ``pretraining + conventional training'' optimization strategy.
However, these methods only apply panoramas to the satellite-based CVGL task.
Leveraging the information-rich panoramas on the OSM-based CVGL task remains unexplored.
In this work, we apply $360${\textdegree}-FoV images to the OSM-based MCVGL task to improve localization accuracy.

\section{Methodology}
\label{sec:method}

\begin{table}[!t]
\begin{center}
\caption{Data distribution of the CV-RHO dataset.}
\vskip -1ex
\label{tab:data-stats}
\setlength{\tabcolsep}{3mm}
\resizebox{\columnwidth}{!}{
\renewcommand{\arraystretch}{1.4}
    \begin{tabular}{lcccc}
    \toprule[1.5pt]
    \textbf{Country}                 & \textbf{City}            & \textbf{360\textdegree-FoV Pan.} & \textbf{120\textdegree-FoV Pin.} & \textbf{OSM} \\
    \midrule
    \multirow{4}{*}{USA}    & Chicago         & 12.4k           & 37.2k   & 12.4k \\
                            & Detroit         & 39.4k           & 118.1k  & 39.4k \\
                            & San Francisco   & 15.5k           & 46.5k  & 15.5k \\
                            & Washington      & 12.5k           & 37.5k  & 12.5k \\
    \midrule
    \multirow{2}{*}{France} & Montrouge       & 9.5k            & 28.5k  & 9.5k \\
                            & Toulouse        & 12.4k           & 37.1k  & 12.4k \\ 
    \midrule
    Germany                 & Berlin          & 12.4k           & 37.1k  & 12.4k \\
    \midrule
    \rowcolor{gray!10}\multicolumn{2}{l}{Total}                 & 114.0k          & 342.1k & 114.0k \\
    \rowcolor{gray!15}\multicolumn{2}{l}{8 variations in total} & \textbf{912.3k}          & \textbf{2.7M}   & \textbf{912.3k} \\
    \bottomrule[1.5pt]
    \end{tabular}
}
\end{center}
\vskip -2ex
\end{table}
\subsection{OSM-Based MCVGL}
\label{sec:MCVGL}
\noindent\textbf{Definition}.
For a single image $\mathbf{I}$ with coarse position prior $\boldsymbol{x}_{prior}$ and known camera calibration, we aim to query the precise camera pose by matching the ground image and OSM.
As OSM mainly consists of 2D geospatial data, it is reasonable to reduce the normal 6-DoF pose estimation to the 3-DoF pose $\boldsymbol{\xi}=\left(u,v,\theta\right)$, where $\left(u,v\right)$ stands for position and $\theta \in \left ( -180^{\circ},180^{\circ}  \right ]$ is for orientation.

\noindent\textbf{Conventions}.
We assume the direction of gravity is known, which can be assessed via an inertial unit embedded in camera devices.
We consider a topocentric coordinate system, with the x-y-z axes corresponding to the East-North-vertical directions.
For the input image $\textbf{I}$, the roll and pitch angles are rectified to zero using the known gravity direction, and the principal axis is ensured to be horizontal.
The OSM is queried as a square area centered around $\boldsymbol{x}_{prior}$, whose coarse extent decides how large the extracted map is.

\begin{table}[!t]
\begin{center}
\vskip -1ex
\caption{Comparison between our CV-RHO dataset and other widely-used CVGL datasets.}
\vskip -1ex
\label{tab:datasets-compare}
\setlength{\tabcolsep}{2mm}
\resizebox{\columnwidth}{!}{
\renewcommand{\arraystretch}{1.2}
    \begin{tabular}{lcccccc}
    \toprule[1.5pt]
    \textbf{Dataset} & \textbf{Pin.}      & \textbf{Pan.}     & \textbf{Satellite}   & \textbf{OSM}       & \textbf{Weather noise} & \textbf{Sensor noise} \\
    \midrule
    CVUSA~\cite{workman2015cvusa}    & \xmark     & \cmark  & \cmark & \xmark     & \xmark                      & \xmark                     \\
    CVACT~\cite{liu2019cvact}    & \xmark     & \cmark  & \cmark & \xmark     & \xmark                      & \xmark                     \\
    VIGOR~\cite{zhu2021vigor}    & \xmark     & \cmark  & \cmark & \xmark     & \xmark                      & \xmark                     \\
    CVGlobal~\cite{ye2024cvglobal} & \xmark     & \cmark & \cmark & \cmark & \xmark                      & \xmark                     \\
    KITTI~\cite{geiger2012kitti}    & \cmark & \xmark  & \cmark    & \cmark & \xmark                      & \xmark                     \\
    MGL~\cite{sarlin2023orienternet}      & \cmark & \xmark  & \xmark   & \cmark & \xmark                      & \xmark                     \\
    \rowcolor{gray!15}\textbf{CV-RHO} (ours)   & \cmark & \cmark  & \xmark & \cmark & \cmark                  & \cmark                 \\
    \bottomrule[1.5pt]
    \end{tabular}
}
\end{center}
\vskip -2ex
\end{table}

\begin{figure}[!t]
    \centering
    \includegraphics[width=\columnwidth]{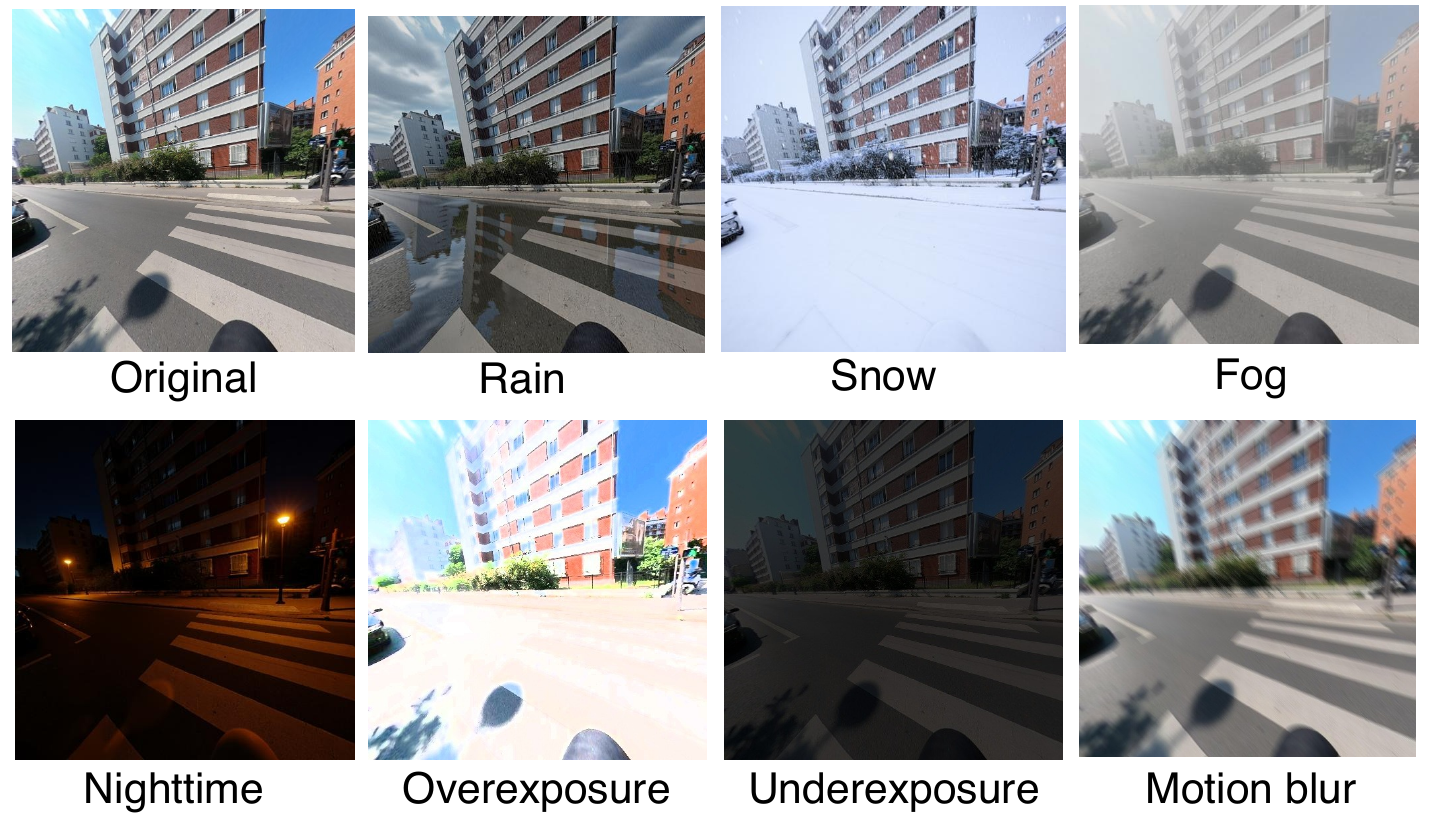}
    \vskip -1ex
    \caption{Data samples of our CV-RHO dataset.}
    \label{fig:data_samples}
    \vskip -2ex
\end{figure}

\begin{figure*}[!h]
    \centering
    \includegraphics[width=0.99\linewidth]{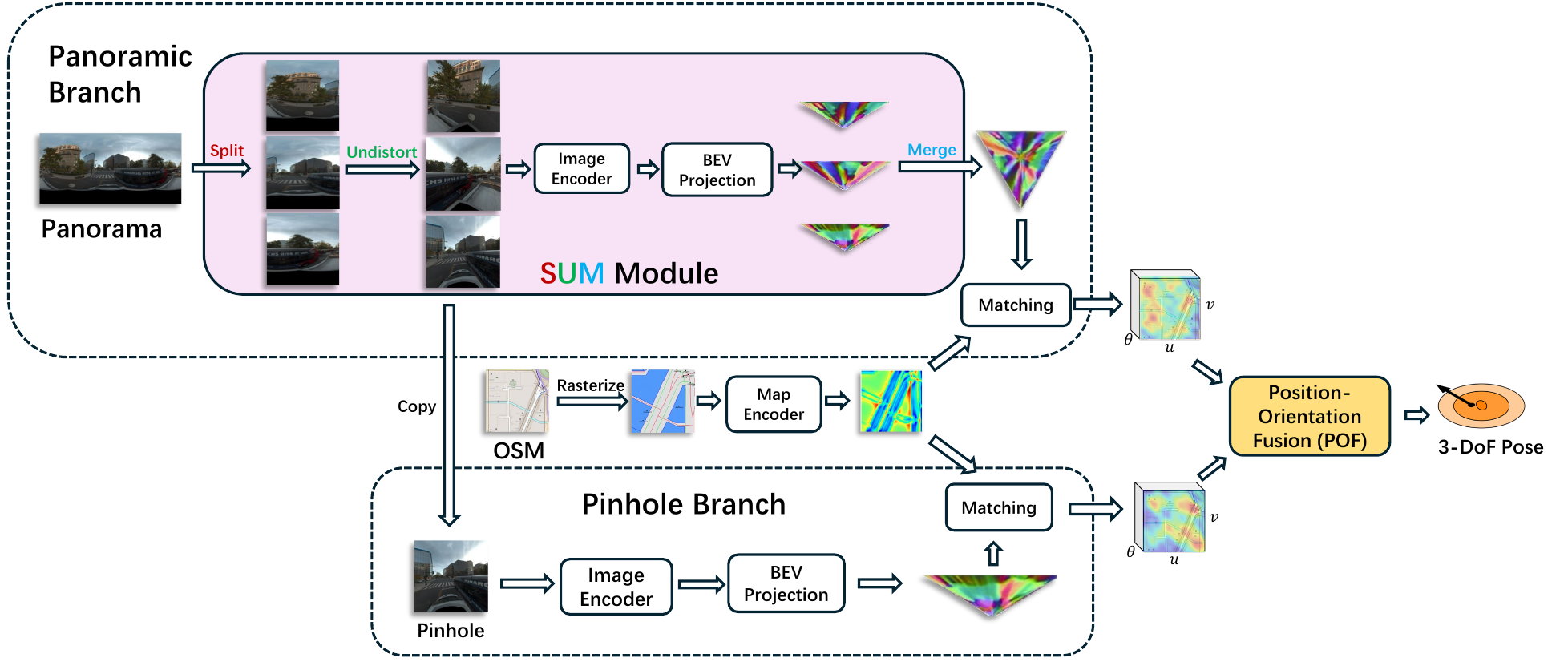}
    \caption{Overview of the RHO model in a two-branch Pin-Pan architecture. Given the current panorama as input, the panoramic branch with Split-Undistort-Merge (SUM) module produces the 360\textdegree-FoV feature, while the pinhole branch focuses on local dense feature in 120\textdegree~FoV. Both generated features are matched with the OSM feature map and then fused by the proposed Position-Orientation Fusion.}
    \label{fig:model-struc}
\end{figure*}
\subsection{CV-RHO Dataset Collection}
\label{sec:cv-rho}
We collect all panoramas and the corresponding metadata using the Mapillary platform.
Given the metadata, we also collect the corresponding OSM tiles.
Regarding varying weather and lighting conditions, FLUX.1 Kontext~\cite{labs2025flux} is applied for image generation under rain, snow, fog, and night, which are representative of common scenarios in MCVGL real-world applications.
The whole image generation process takes $30.7$k A100 GPU hours.
The prompts and parameters used for generating each weather condition are provided in our supplementary material.
For perturbations like motion blur, overexposure, and underexposure, OpenCV~\cite{opencv_library} is used to introduce noise effects into the original images.
These three variations involve changes in brightness and exposure duration, which effectively replicate real-world conditions.
Fig.~\ref{fig:data_samples} presents data samples of the CV-RHO dataset under varying conditions.
It is worth noting that all generated images are structurally consistent with the original image.
Table~\ref{tab:data-stats} and \ref{tab:datasets-compare} showcase the data distribution of the CV-RHO dataset and the comparison with other widely-used CVGL datasets, respectively. We split the dataset into train and test sets with the ratio of $8:2$.

To assess the generalization ability of models, we additionally collect a test set with $10$k images in Mount Vernon, USA.
This cross-region dataset is used for evaluating the cross-region localization ability instead of the same-region localization ability using the test set of the CV-RHO dataset.
The results are listed in Table~\ref{tab:cross-region} and will be discussed in Section~\ref{sec:experiments}.

To assess the quality of the generated data, we additionally collect a test set with $5$k images and evaluate models using this test set in a zero-shot manner.
This Sim2Real test set contains photos captured under real-world weather perturbations.
The results are listed in Table~\ref{tab:sim2real} and will be discussed in Section~\ref{sec:experiments}.

\subsection{RHO Model}
\label{subsec:rho}
\noindent\textbf{RHO Model Structure}.
Fig.~\ref{fig:model-struc} presents the overview of our RHO model.
RHO adopts a two-branch Pin–Pan architecture. 
The Shannon Entropy of an image is defined as:
\begin{align}
    H = - \sum_{i} p_i \log_2 p_i,
\end{align}
where $p_i$ stands for the occurring probability of the $i$-th pixel.
Compared to pinhole images, a panorama provides richer visual information for the estimation of the camera position due to the larger FoV, resulting in larger Shannon Entropy.
However, referring to the prediction of the orientation, the panorama is hard to bring additional performance gain since the Shannon Entropy remains the same when the camera rotates by an angle.
In this case, the pinhole image showcases its potential in predicting the orientation, while the Shannon Entropy differs after the camera's rotation.

The panoramic branch processes a panoramic image to produce a BEV feature map covering a full 360\textdegree~FoV, while the pinhole branch takes a pinhole image as input and generates a BEV feature map with a 120\textdegree~FoV.
The OSM tile is first rasterized and then fed into a map encoder, resulting in an OSM feature map.
The two BEV feature maps are then respectively matched with the OSM feature map, outputting two probability volumes.
Each volume has three dimensions, namely $u$, $v$ and $\theta$ representing width, height and the number of rotation angles, respectively.
These two volumes are fused by the Position-Orientation Fusion (POF) module, and the RHO model finally estimates the 3-DoF camera pose.

\noindent \textbf{SUM Module}.
A panoramic ground image is split into $3$ views.
Each view is a $120^\circ$-FoV pinhole image representing the yaw angle ranging in $\left [ 0^{\circ}, 120^{\circ} \right )$, $\left [ 120^{\circ}, 240^{\circ} \right )$, and $\left [ 240^{\circ}, 360^{\circ} \right )$, respectively.
After the splitting, all pinhole images are then undistorted to remove the panoramic deformation.
Each pinhole image $\mathbf{I}_{i}, i \in \left\{1,2,3\right\}$ is encoded via $\Phi_{image}$ and $\Phi_{bev}$ forming a BEV feature map $\mathbf{X}_{i},i \in \left\{1,2,3\right\}$.
All $\mathbf{X}_{i}$ are then merged, resulting in a panoramic BEV feature map $\mathbf{X}_{pan}$.

\noindent \textbf{Map Encoding.} The 2D map with areas, lines, and points classes is first rasterized as a $3$-channel image. Each semantic class is then mapped to a unique semantic vector. We employ the $\Phi_{map}$ to encode the rasterized map into a neural map $\mathbf{M}$, which will be utilized in the map-BEV matching.

\begin{figure*}
    \centering
    \includegraphics[width=1.0\linewidth]{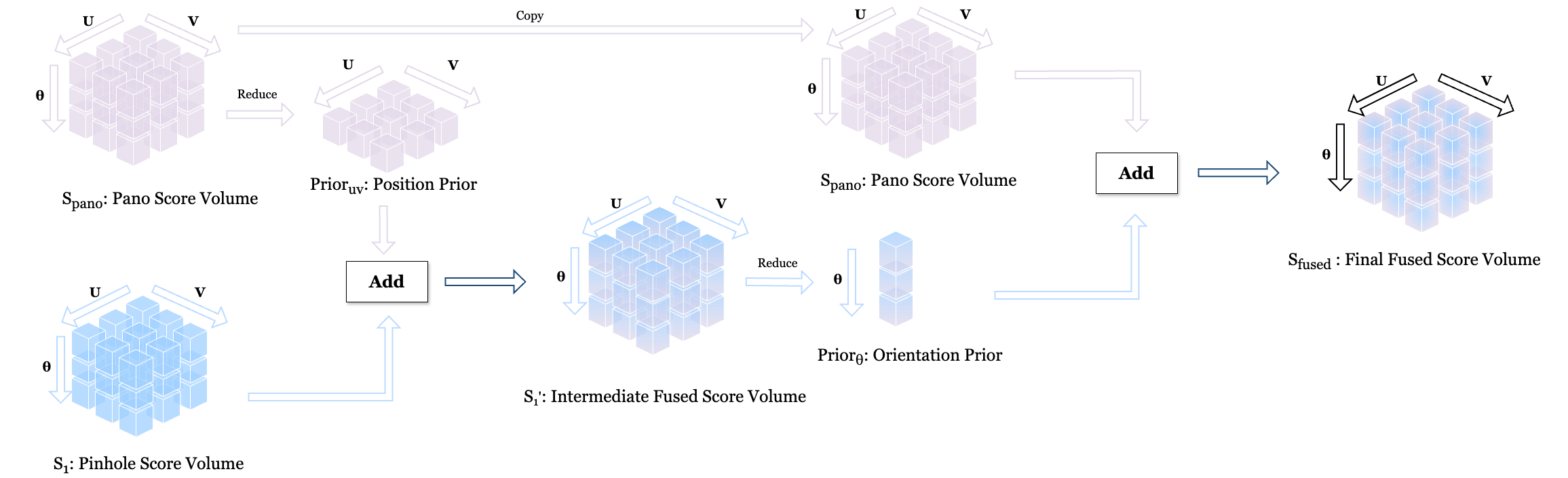}
    \caption{Position-Orientation Fusion module of the RHO model. Given both 360\textdegree-FoV and 120\textdegree-FoV volumes, POF combines the pinhole volume $\mathbf{S}_{1}$ with $Prior_{uv}$, and then the panoramic volume $\mathbf{S}_{pano}$ with $Prior_{\theta}$.}
    \label{fig:cross-fusion}
\end{figure*}

\noindent\textbf{POF Module}.
By exhaustively matching the neural map $\mathbf{M}$ and panoramic BEV feature map $\mathbf{X}_{pan}$, we yield a score volume $\mathbf{S}_{pano}$. Meanwhile, we calculate the score volume $\mathbf{S}_{1}$ by matching the neural map $\mathbf{M}$ and the pinhole BEV feature map $\mathbf{X}_{1}$ of the front view, representing the yaw angle ranging in $\left [ 0^{\circ}, 120^{\circ} \right )$.
Note that $\mathbf{X}_{2}$ and $\mathbf{X}_{3}$ can alternatively serve as inputs to the pinhole branch, provided that the ground-truth camera orientation is adjusted by $+120^{\circ}$ and $+240^{\circ}$, respectively.
$\mathbf{S}_{pano}$ contains accurate prediction of position, while $\mathbf{S}_{1}$ focuses more on the orientation. In order to leverage the advantages of both $\mathbf{S}_{pano}$ and $\mathbf{S}_{1}$, we propose a POF module, a two-stage fusion process shown in Fig.~\ref{fig:cross-fusion}.
In the first stage, for location $\boldsymbol{x}=(u,v)$, and camera pose $\boldsymbol{\xi}=(\boldsymbol{x},\theta)$, we calculate a robust 2D spatial prior by marginalizing the panorama log-probability volume over orientation dimension $\theta$ using LogSumExp (LSE). The 2D prior highlights likely locations regardless of orientation, expressed in Eq.~\ref{eq:stage1-1} and Eq.~\ref{eq:stage1-2}.
\begin{equation}
    \log P_{pano}(\boldsymbol{\xi})=\mathrm{LogSoftmax}(\boldsymbol{S}_{pano}(\boldsymbol{\xi}))
    \label{eq:stage1-1}
\end{equation}
\vspace{-1em}
\begin{equation}
    \log P_{uv\_prior}(\boldsymbol{x})=\mathrm{LSE}_{\theta}(\log P_{pano}(\boldsymbol{\xi}))
    \label{eq:stage1-2}
\end{equation}
The pinhole log-probability volume $\log P_{1}(\boldsymbol{\xi})$ is then re-weighted by the spatial prior, expressed in Eq.~\ref{eq:stage1-3} and Eq.~\ref{eq:stage1-4}, where $\alpha$ is a learnable hyper-parameter.
\begin{equation}
    \log P_{1}(\boldsymbol{\xi})=\mathrm{LogSoftmax}(\boldsymbol{S}_{1}(\boldsymbol{\xi}))
    \label{eq:stage1-3}
\end{equation}
\vspace{-1.8em}
\begin{equation}
    \log \boldsymbol{S}_{1}^{'}(\boldsymbol{\xi})=(1-\alpha) \cdot \log P_{1}(\boldsymbol{\xi}) + \alpha \cdot \log P_{uv\_prior}(\boldsymbol{x})
    \label{eq:stage1-4}
\end{equation}
In the second stage, a sharp orientation prior is obtained by normalizing $\mathbf{S}_{1}^{'}$ and marginalizing over the spatial dimensions $\boldsymbol{x}$.
The panorama score volume $\mathbf{S}_{pano}$ is fused with the orientation prior $\log P_{\theta\_prior}(\theta)$ as the final fused score volume $\mathbf{S}_{fused}$, expressed in Eq.~\ref{eq:stage2-1}, Eq.~\ref{eq:stage2-2} and Eq.~\ref{eq:stage2-3}, where $\beta$ is a learnable hyper-parameter.
\begin{equation}
    \log P_{1}^{'}(\boldsymbol{\xi})=\mathrm{LogSoftmax}(\boldsymbol{S}_{1}^{'}(\boldsymbol{\xi}))
    \label{eq:stage2-1}
\end{equation}
\vspace{-1.5em}
\begin{equation}
    \log P_{\theta\_prior}(\theta)=\mathrm{LSE}_{\boldsymbol{x}}(\log P_{1}^{'}(\boldsymbol{\xi}))
    \label{eq:stage2-2}
\end{equation}
\vspace{-1.8em}
\begin{equation}
\resizebox{0.9\linewidth}{!}{$
\log \boldsymbol{S}_{fused}(\boldsymbol{\xi}) = (1-\beta) \cdot \log P_{pano}(\boldsymbol{\xi}) + \beta \cdot \log P_{\theta\_prior}(\theta)
$}
\label{eq:stage2-3}
\end{equation}
The final score volume $\mathbf{S}_{fused}$ is then used for determining the final camera pose.

\section{Experiments}
\label{sec:experiments}
\subsection{Implementation Details}
\label{sec:implementation}
We conduct experiments on our large-scale CV-RHO dataset.
We use $12$ NVIDIA A100 GPUs to train the RHO model.
The number of rotations is $64$ during training and $256$ in evaluation.
The batch size and learning rate are set to $36$ and $2e^{-5}$, respectively.
We choose ReduceLROnPlateau as the learning rate scheduler to manage the learning rate and Adam optimizer for training.
With the settings above, the model achieves the best result between epochs $2$ and $4$, step $20$k to $40$k approximately.

\subsection{Quantitative Results}
\label{sec:quantitative}
\begin{table}[!t]
\centering
\caption{Results on clean CV-RHO dataset. Position Recall (PR) and Orientation Recall (OR) are in percentages.}
\label{tab:exp1}
\setlength{\tabcolsep}{2mm}
\resizebox{\columnwidth}{!}{%
\begin{tabular}{lccccccc}
\toprule[1.5pt]
\multirow{2}{*}{\textbf{Approach}} &
  \multirow{2}{*}{\textbf{FoV}} &
  \multicolumn{3}{c}{\textbf{PR @ Xm~$\uparrow$}} &
  \multicolumn{3}{c}{\textbf{OR @ X°~$\uparrow$}} \\ \cline{3-8} 
                             &                             & 1m             & 3m             & 5m             & 1°             & 3°             & 5°             \\ \midrule
OrienterNet~\cite{sarlin2023orienternet} & 90\textdegree   & 18.02          & 58.37          & 71.04          & 27.72          & 63.86          & 77.50           \\
OrienterNet~\cite{sarlin2023orienternet}              & 120\textdegree            & 21.83          & 66.16          & 78.03          & 35.02          & 74.89          & 85.62          \\
OrienterNet~\cite{sarlin2023orienternet}              & 360\textdegree            & 3.79           & 19.35          & 28.78          & 10.29          & 28.43          & 36.87          \\
\rowcolor{gray!15}\textbf{RHO} (ours)                          & 360\textdegree                  & \textbf{24.59} & \textbf{73.55} & \textbf{84.36} & \textbf{43.46} & \textbf{83.61} & \textbf{90.44} \\
\bottomrule[1.5pt]
\end{tabular}%
}
\end{table}
\begin{table}[!t]
\centering
\caption{Results on CV-RHO dataset under varying conditions and noises. Position Recall (PR) and Orientation Recall (OR) are in percentages. NT: nighttime; OE: over-exposure; UE: under-exposure; MB: motion blur; ANV: all noisy variations; AV: all variations; Average Degradation is overall performance drop.}
\label{tab:exp1.2}
\setlength{\tabcolsep}{1.5mm}
\resizebox{\columnwidth}{!}{%
\begin{tabular}{lcc|cccccc}
\toprule[1.5pt]
\multirow{2}{*}{\textbf{Method}} &
  \multicolumn{2}{c|}{\textbf{Variation}} &
  \multicolumn{3}{c}{\textbf{PR @ Xm}~$\uparrow$} &
  \multicolumn{3}{c}{\textbf{OR @ X°}~$\uparrow$} \\ \cline{2-9} 
 & Train      & Eval               & 1m    & 3m    & 5m    & 1°    & 3°    & 5°    \\
 \midrule
\multirow{11}{*}{\rot{OrienterNet~\cite{sarlin2023orienternet}}} &
  {Clean} &
  {Clean} &
  {21.83} &
  {66.16} &
  {78.03} &
  {35.02} &
  {74.89} &
  {85.62} \\
 &     {Clean}          & Rain                     & 13.98 & 52.98 & 68.12 & 25.51 & 60.49 & 74.68 \\
 &     {Clean}          & NT                & 11.44 & 44.55 & 57.77 & 20.71 & 51.13 & 64.84 \\
 &     {Clean}          & Fog                      & 16.26 & 54.52 & 67.12 & 26.81 & 62.27 & 75.02 \\
 &     {Clean}          & Snow                     & 14.29 & 50.02 & 63.23 & 24.30  & 57.52 & 70.68 \\
 &     {Clean}          & OE            & 16.36 & 57.57 & 70.86 & 28.90  & 66.00    & 78.55 \\
 &     {Clean}          & UE           & 19.74 & 63.25 & 75.65 & 32.88 & 71.90 & 83.35 \\
 &     {Clean}          & MB              & 4.90   & 20.98 & 30.30  & 9.15  & 24.72 & 35.05 \\
 &     {Clean}          & ANV     & 12.41 & 46.43 & 60.03 & 23.51 & 55.06 & 67.44 \\
 &     {Clean}          & AV           & 13.15 & 48.46 & 62.17 & 24.56 & 57.25 & 69.63 \\
 \multicolumn{3}{r|}{\textbf{Average Degradation}}
   &
  \textcolor{red}{\textbf{-8.22}} &
  \textcolor{red}{\textbf{-17.41}} &
  \textcolor{red}{\textbf{-16.34}} &
  \textcolor{red}{\textbf{-10.98}} &
  \textcolor{red}{\textbf{-18.63}} &
  \textcolor{red}{\textbf{-16.82}} \\
  \midrule
\multirow{11}{*}{\rot{\textbf{RHO} (ours)}} &
  {Clean} &
  {Clean} &
  {24.59} &
  {73.55} &
  {84.36} &
  {43.46} &
  {83.61} &
  {90.44} \\
 &      {Clean}         & Rain                     & 20.91 & 68.48 & 80.77 & 37.58 & 78.25 & 87.19 \\
 &      {Clean}         & NT                & 17.46 & 59.83 & 72.48 & 30.74 & 68.37 & 79.51 \\
 &      {Clean}         & Fog                      & 20.27 & 64.42 & 75.93 & 35.23 & 72.63 & 82.23 \\
 &      {Clean}         & Snow                     & 18.95 & 62.33 & 74.14 & 32.35 & 70.45 & 80.73 \\
 &      {Clean}         & OE            & 20.23 & 67.73 & 79.46 & 38.17 & 78.12 & 86.51 \\
 &      {Clean}         & UE           & 23.49 & 71.95 & 83.17 & 41.93 & 82.25 & 89.37 \\
 &      {Clean}         & MB              & 10.15 & 37.50  & 49.69 & 18.15 & 45.06 & 57.82 \\
 &      {Clean}         & ANV     & 17.67 & 60.08 & 73.01 & 32.99 & 70.51 & 80.52 \\
 &      {Clean}         & AV           & 18.48 & 61.51 & 74.38 & 34.47 & 72.01 & 81.77 \\
 \multicolumn{3}{r|}{\textbf{Average Degradation}} 
 &\cellcolor{gray!15}
  \textcolor{red}{\textbf{-5.97}} &\cellcolor{gray!15}
  \textcolor{red}{\textbf{-12.01}} &\cellcolor{gray!15}
  \textcolor{red}{\textbf{-10.69}} &\cellcolor{gray!15}
  \textcolor{red}{\textbf{-9.95}} &\cellcolor{gray!15}
  \textcolor{red}{\textbf{-12.76}} &\cellcolor{gray!15}
  \textcolor{red}{\textbf{-9.81}} \\
  \midrule
  \midrule
\multirow{10}{*}{\rot{OrienterNet~\cite{sarlin2023orienternet}}} &
  {Clean} &
  {Clean} &
  {21.83} &
  {66.16} &
  {78.03} &
  {35.02} &
  {74.89} &
  {85.62} \\
  &{Rain} & {Rain} &
  20.48 &
  64.36 &
  76.57 &
  33.89 &
  73.41 &
  84.49 \\
 & {NT} & {NT}          & 19.26 & 60.76 & 73.04 & 31.75 & 70.06 & 81.42 \\
 & {Fog} & {Fog}                 & 20.32 & 63.39 & 75.53 & 33.26 & 72.42 & 83.43 \\
 & {Snow}  & {Snow}               & 20.29 & 63.56 & 75.68 & 33.57 & 72.54 & 83.72 \\
 & {OE}  & {OE}       & 20.32 & 63.81 & 76.15 & 33.09 & 72.58 & 83.88 \\
 & {UE} & {UE}      & 21.36 & 65.16 & 77.41 & 34.80  & 74.31 & 85.02 \\
 & {MB}  & {MB}        & 19.78 & 62.29 & 74.56 & 32.68 & 71.33 & 82.63 \\
 & {ANV} & {ANV} & 18.13 & 60.36 & 73.70  & 32.89 & 71.38 & 82.24 \\
 & {AV} & {AV}      & 18.14 & 60.72 & 74.21 & 32.83 & 71.48 & 82.34 \\
 \multicolumn{3}{r|}{\textbf{Average Degradation}}&
  \textcolor{red}{\textbf{-2.04}} &
  \textcolor{red}{\textbf{-3.45}} &
  \textcolor{red}{\textbf{-2.82}} &
  \textcolor{red}{\textbf{-1.82}} &
  \textcolor{red}{\textbf{-2.72}} &
  \textcolor{red}{\textbf{-2.38}} \\
  \midrule
\multirow{10}{*}{\rot{\textbf{RHO} (ours)}} &
  {Clean} &
  {Clean} &
  {24.59} &
  {73.55} &
  {84.36} &
  {43.46} &
  {83.61} &
  {90.44} \\
  &{Rain} &  {Rain} &
  24.8 &
  73.91 &
  84.55 &
  42.28 &
  83.35 &
  90.63 \\
 & {NT} & {NT}            & 24.46 & 72.04 & 83.01 & 41.23 & 81.53 & 89.44 \\
 & {Fog} & {Fog}                  & 25.34 & 73.12 & 83.93 & 41.89 & 82.72 & 90.35 \\
 & {Snow}  & {Snow}                 & 24.75 & 72.68 & 83.56 & 43.09 & 82.72 & 89.82 \\
 & {OE} & {OE}        & 24.94 & 73.20  & 83.80  & 41.68 & 82.05 & 89.83 \\
 & {UE}  & {UE}      & 25.91 & 74.42 & 84.72 & 43.23 & 83.61 & 90.67 \\
 & {MB} & {MB}          & 25.39 & 73.45 & 84.07 & 42.13 & 82.63 & 90.15 \\
 & {ANV} & {ANV} & 23.02 & 72.11 & 83.58 & 42.52 & 82.43 & 90.00    \\
 & {AV} & {AV}       & 22.98 & 71.58 & 83.32 & 43.22 & 82.50  & 89.91 \\
 \multicolumn{3}{r|}{\textbf{Average Degradation}} &\cellcolor{gray!15}
  \textcolor{red}{\textbf{+0.03}} &\cellcolor{gray!15}
  \textcolor{red}{\textbf{-0.60}} &\cellcolor{gray!15}
  \textcolor{red}{\textbf{-0.52}} &\cellcolor{gray!15}
  \textcolor{red}{\textbf{-1.10}} &\cellcolor{gray!15}
  \textcolor{red}{\textbf{-0.99}} &\cellcolor{gray!15}
  \textcolor{red}{\textbf{-0.35}} \\
  \bottomrule[1.5pt]
\end{tabular}
}
\end{table}
\noindent\textbf{Results on Clean CV-RHO}
Table~\ref{tab:exp1} showcases the results on the clean CV-RHO dataset, namely using clean, daytime ground images for training.
When trained with $120^{\circ}$-FoV pinhole images, OrienterNet achieves better recall than when trained with $90^\circ$-FoV images on CV-RHO.
This shows that a broader FoV boosts the model performance.
However, the $360^{\circ}$-FoV OrienterNet has the worst results due to the lack of the ability to handle panoramic distortion.
Our model RHO outperforms other baselines with significant performance gains of $2.76\%${/}$7.39\%${/}$6.33\%$ on PR@$1{/}3{/}5$m and $8.44\%${/}$8.72\%${/}$4.82\%$ on OR@$1{/}3{/}5^\circ$.
This performance boost proves that RHO achieves the best results due to the reasonable model design rather than purely leveraging large-FoV images.

\begin{table}[!t]
\centering
\caption{Results of cross-region evaluation. OrienterNet and our model RHO are both trained on the CV-RHO under various conditions and evaluated on the new city dataset. Position Recall (PR) and Orientation Recall (OR) are in percentages. MV: MountVernon. NT: nighttime; OE: over-exposure; UE: under-exposure; MB: motion blur; ANV: all noisy variations; AV: all variations.}
\label{tab:cross-region}
\setlength{\tabcolsep}{1.2mm}
\begin{adjustbox}{width=\linewidth}
\begin{tabular}{lcccccccc}
\toprule[1.5pt]
\multirow{2}{*}{\textbf{Method}} &
  \multirow{2}{*}{\textbf{Variation}} &
  \multirow{2}{*}{\textbf{Dataset}} &
  \multicolumn{3}{c}{\textbf{PR @ Xm}~$\uparrow$} &
  \multicolumn{3}{c}{\textbf{OR @ X°}~$\uparrow$} \\ \cline{4-9} 
                  &                      &                   & 1m    & 3m    & 5m    & 1°    & 3°    & 5°    \\ \midrule
OrienterNet~\cite{sarlin2023orienternet} & \multirow{4}{*}{Clean} & \multirow{2}{*}{MV}       & 16.36 & 54.17 & 62.82 & 25.62 & 58.56 & 69.80  \\
\textbf{RHO} (ours)         &                      &                   & 26.24 & 67.38 & 73.79 & 38.00    & 73.35 & 80.20  \\ \cdashline{4-9} 
OrienterNet~\cite{sarlin2023orienternet} &                      & \multirow{2}{*}{CV-RHO} & 21.83 & 66.16 & 78.03 & 35.02 & 74.89 & 85.62 \\
\textbf{RHO} (ours)         &                      &                   & 24.59 & 73.55 & 84.36 & 43.46 & 83.61 & 90.44 \\
\midrule
OrienterNet~\cite{sarlin2023orienternet} & \multirow{4}{*}{Rain} & \multirow{2}{*}{MV}        & 15.94 & 53.03 & 61.75 & 25.34 & 57.82 & 69.40  \\
\textbf{RHO} (ours)         &                      &                   & 26.37 & 66.75 & 73.27 & 37.79 & 72.67 & 78.90  \\ \cdashline{4-9} 
OrienterNet~\cite{sarlin2023orienternet} &                      & \multirow{2}{*}{CV-RHO} & 20.48 & 64.36 & 76.57 & 33.89 & 73.41 & 84.49 \\
\textbf{RHO} (ours)         &                      &                   & 24.80  & 73.91 & 84.55 & 42.28 & 83.35 & 90.63 \\
\midrule
OrienterNet~\cite{sarlin2023orienternet} & \multirow{4}{*}{NT} & \multirow{2}{*}{MV}          & 14.14 & 48.59 & 57.83 & 22.71 & 54.33 & 65.67 \\
\textbf{RHO} (ours)         &                      &                   & 25.10  & 65.27 & 71.30  & 36.13 & 70.21 & 77.76 \\ \cdashline{4-9} 
OrienterNet~\cite{sarlin2023orienternet} &                      & \multirow{2}{*}{CV-RHO} & 19.26 & 60.76 & 73.04 & 31.75 & 70.06 & 81.42 \\
\textbf{RHO} (ours)         &                      &                   & 24.46 & 72.04 & 83.01 & 41.23 & 81.53 & 89.44 \\
\midrule
OrienterNet~\cite{sarlin2023orienternet} & \multirow{4}{*}{Fog} & \multirow{2}{*}{MV}         & 15.43 & 51.55 & 60.78 & 24.13 & 56.54 & 68.22 \\
\textbf{RHO} (ours)         &                      &                   & 23.83 & 65.07 & 72.49 & 36.41 & 71.76 & 78.15 \\ \cdashline{4-9} 
OrienterNet~\cite{sarlin2023orienternet} &                      & \multirow{2}{*}{CV-RHO} & 20.32 & 63.39 & 75.53 & 33.26 & 72.42 & 83.43 \\
\textbf{RHO} (ours)         &                      &                   & 25.34 & 73.12 & 83.93 & 41.89 & 82.72 & 90.35 \\
\midrule
OrienterNet~\cite{sarlin2023orienternet} & \multirow{4}{*}{Snow} & \multirow{2}{*}{MV}        & 13.71 & 48.60  & 57.93 & 23.74 & 54.58 & 66.29 \\
\textbf{RHO} (ours)         &                      &                   & 23.62 & 64.21 & 71.58 & 37.50  & 71.35 & 78.22 \\ \cdashline{4-9} 
OrienterNet~\cite{sarlin2023orienternet} &                      & \multirow{2}{*}{CV-RHO} & 20.29 & 63.56 & 75.68 & 33.57 & 72.54 & 83.72 \\
\textbf{RHO} (ours)         &                      &                   & 24.75 & 72.68 & 83.56 & 43.09 & 82.72 & 89.82 \\
\midrule
OrienterNet~\cite{sarlin2023orienternet} & \multirow{4}{*}{OE} & \multirow{2}{*}{MV}          & 14.92 & 51.37 & 60.66 & 24.60  & 56.89 & 68.36 \\
\textbf{RHO} (ours)         &                      &                   & 24.42 & 62.78 & 69.06 & 35.38 & 69.35 & 75.86 \\ \cdashline{4-9} 
OrienterNet~\cite{sarlin2023orienternet} &                      & \multirow{2}{*}{CV-RHO} & 20.32 & 63.81 & 76.15 & 33.09 & 72.58 & 83.88 \\
\textbf{RHO} (ours)         &                      &                   & 24.94 & 73.20  & 83.80  & 41.68 & 82.05 & 89.83 \\
\midrule
OrienterNet~\cite{sarlin2023orienternet} & \multirow{4}{*}{UE} & \multirow{2}{*}{MV}          & 16.33 & 53.48 & 62.60  & 25.85 & 59.17 & 70.72 \\
\textbf{RHO} (ours)         &                      &                   & 25.51 & 64.78 & 70.98 & 35.95 & 70.85 & 77.68 \\ \cdashline{4-9} 
OrienterNet~\cite{sarlin2023orienternet} &                      & \multirow{2}{*}{CV-RHO} & 21.36 & 65.16 & 77.41 & 34.80  & 74.31 & 85.02 \\
\textbf{RHO} (ours)         &                      &                   & 25.91 & 74.42 & 84.72 & 43.23 & 83.61 & 90.67 \\
\midrule
OrienterNet~\cite{sarlin2023orienternet} & \multirow{4}{*}{MB} & \multirow{2}{*}{MV}          & 13.42 & 48.30  & 57.82 & 23.28 & 53.85 & 66.17 \\
\textbf{RHO} (ours)         &                      &                   & 24.60  & 64.31 & 71.71 & 35.56 & 70.41 & 77.60  \\ \cdashline{4-9} 
OrienterNet~\cite{sarlin2023orienternet} &                      & \multirow{2}{*}{CV-RHO} & 19.78 & 62.29 & 74.56 & 32.68 & 71.33 & 82.63 \\
\textbf{RHO} (ours)         &                      &                   & 25.39 & 73.45 & 84.07 & 42.13 & 82.63 & 90.15 \\
\midrule
OrienterNet~\cite{sarlin2023orienternet} &
  \multirow{4}{*}{ANV} &
  \multirow{2}{*}{MV} &
  14.58 &
  49.77 &
  58.69 &
  24.20 &
  55.56 &
  66.81 \\
\textbf{RHO} (ours)         &                      &                   & 24.96 & 65.32 & 71.58 & 36.53 & 72.03 & 78.20  \\ \cdashline{4-9} 
OrienterNet~\cite{sarlin2023orienternet} &                      & \multirow{2}{*}{CV-RHO} & 18.13 & 60.36 & 73.70  & 32.89 & 71.38 & 82.24 \\
\textbf{RHO} (ours)         &                      &                   & 23.02 & 72.11 & 83.58 & 42.52 & 82.43 & 90.00    \\
\midrule
OrienterNet~\cite{sarlin2023orienternet} & \multirow{4}{*}{AV} & \multirow{2}{*}{MV}  & 14.51 & 50.24 & 59.29 & 24.77 & 56.58 & 67.31 \\
\textbf{RHO} (ours)         &                      &                   & 25.10  & 65.78 & 72.13 & 37.65 & 72.72 & 78.48 \\ \cdashline{4-9} 
OrienterNet~\cite{sarlin2023orienternet} &                      & \multirow{2}{*}{CV-RHO} & 18.14 & 60.72 & 74.21 & 32.83 & 71.48 & 82.34 \\
\textbf{RHO} (ours)         &                      &                   & 22.98 & 71.58 & 83.32 & 43.22 & 82.50  & 89.91 \\
\bottomrule[1.5pt]
\end{tabular}
\end{adjustbox}
\end{table}
\noindent\textbf{Results on Noisy CV-RHO}.
We first train OrienterNet and RHO using the clean variation of the CV-RHO dataset, then evaluate them on all variations.
The results are reported in the upper part of Table~\ref{tab:exp1.2}.
It can be observed that both OrienterNet and RHO have performance degradation in such a zero-shot setting.
However, the performance drop of RHO is lower than OrienterNet on both positional and orientational recalls.
It proves that RHO is superior on the OSM-based MCVGL task.

We then train and evaluate OrienterNet and RHO using every single variation.
All results are listed in the lower part of Table~\ref{tab:exp1.2}.
Both OrienterNet and RHO are able to handle different weather and lighting conditions, as well as sensor noise, after training on noisy CV-RHO.
Compared with the clean subset, the performance of OrienterNet degrades on other adverse subsets.
However, when comparing the average degradation of OrienterNet between the zero-shot and noisy settings, the performance drop is significantly smaller.
It proves that models trained on our proposed CV-RHO dataset are capable of handling different conditions and perturbations.
CV-RHO can be used to benchmark the robust OSM-based MCVGL task.
As for RHO trained on the noisy CV-RHO dataset, the average degradation is minor compared with the clean subset.
Moreover, RHO consistently outperforms OrienterNet in the noisy setting, showcasing its robustness under adverse conditions.

\begin{table*}[!t]
\centering
\caption{Sim2Real Results. Position Recall (PR) and Orientation Recall (OR) are in percentages. MV: Mount Vernon; OE: over-exposure; UE: under-
exposure; MB: motion blur.} 
\label{tab:sim2real}
\setlength{\tabcolsep}{4mm}
\renewcommand{\arraystretch}{1.1}
\resizebox{\textwidth}{!}{%
\begin{tabular}{lccc|cccccc}
\toprule[1.5pt]
\multirow{2}{*}{\textbf{Method}} &
  \multirow{2}{*}{\textbf{Dataset}} &
  \multirow{2}{*}{\textbf{Variation}} &
  \multirow{2}{*}{\textbf{City}} &
  \multicolumn{3}{c}{\textbf{PR @ Xm}~$\uparrow$} &
  \multicolumn{3}{c}{\textbf{OR @ X°}~$\uparrow$} \\ \cline{5-10} 
                              &                               &                &                               & 1m    & 3m    & 5m    & 1°    & 3°    & 5°    \\ \midrule
\multirow{10}{*}{OrienterNet~\cite{sarlin2023orienternet}} & \multirow{5}{*}{Sim2Real}     & Rain          & Ogori                         & 4.85  & 40.20  & 56.36 & 18.59 & 47.47 & 61.62 \\
                              &                               & Snow          & Detroit                       & 11.16 & 47.07 & 57.54 & 22.59 & 52.13 & 63.95 \\
                              &                               & OE  & Hasselt                       & 3.97  & 29.71 & 47.84 & 17.62 & 47.67 & 63.04 \\
                              &                               & UE & Hasselt                       & 2.07  & 17.10  & 28.67 & 10.19 & 27.29 & 39.03 \\
                              &                               & MB    & Hasselt                       & 3.87  & 28.62 & 45.29 & 15.99 & 46.63 & 62.12 \\ \cline{3-10} 
                              & \multirow{5}{*}{Cross Region} & Rain          & \multirow{5}{*}{MV}  & 15.94 & 53.03 & 61.75 & 25.34 & 57.82 & 69.40  \\
                              &                               & Snow          &                               & 13.71 & 48.6  & 57.93 & 23.74 & 54.58 & 66.29 \\
                              &                               & OE  &                               & 14.92 & 51.37 & 60.66 & 24.6  & 56.89 & 68.36 \\
                              &                               & UE &                               & 16.33 & 53.48 & 62.60  & 25.85 & 59.17 & 70.72 \\
                              &                               & MB    &                               & 13.42 & 48.30  & 57.82 & 23.28 & 53.85 & 66.17 \\ \midrule
\multirow{10}{*}{RHO (ours)}         & \multirow{5}{*}{Sim2Real}     & Rain          & Ogori                         & 3.64  & 54.55 & 68.48 & 32.73 & 69.09 & 81.82 \\
                              &                               & Snow          & Detroit                       & 16.32 & 54.03 & 62.91 & 33.16 & 66.63 & 72.52 \\
                              &                               & OE  & Hasselt                       & 5.18  & 36.79 & 62.69 & 23.83 & 59.07 & 76.17 \\
                              &                               & UE & Hasselt                       & 6.70   & 42.27 & 61.86 & 24.74 & 60.31 & 76.29 \\
                              &                               & MB    & Hasselt                       & 3.54  & 42.93 & 63.13 & 27.27 & 61.11 & 79.29 \\ \cline{3-10} 
                              & \multirow{5}{*}{Cross Region} & Rain          & \multirow{5}{*}{MV} & 26.37 & 66.75 & 73.27 & 37.79 & 72.67 & 78.90  \\
                              &                               & Snow          &                               & 23.62 & 64.21 & 71.58 & 37.50  & 71.35 & 78.22 \\
                              &                               & OE  &                               & 24.42 & 62.78 & 69.06 & 35.38 & 69.35 & 75.86 \\
                              &                               & UE &                               & 25.51 & 64.78 & 70.98 & 35.95 & 70.85 & 77.68 \\
                              &                               & MB    &                               & 24.60  & 64.31 & 71.71 & 35.56 & 70.41 & 77.60  \\
                              \bottomrule[1.5pt]
\end{tabular}
}
\end{table*}

\noindent\textbf{Results of Cross-Region Evaluation}.
Generalization is an important consideration when designing models for the OSM-based MCVGL task.
Previous experiments focus on the same-region evaluation, meaning that models are trained and evaluated using the data from the same city of the CV-RHO dataset.
For the sake of evaluating the cross-region localization ability of the RHO model, we additionally collect a test set with $10$k images in Mount Vernon, USA.
We train the RHO model using the CV-RHO dataset while evaluating the model using the cross-region test set.
The results are reported in Table~\ref{tab:cross-region}.
We also report the same-region results in Table~\ref{tab:cross-region} for comparison.
It can be observed that the domain gap exists.
However, RHO consistently outperforms OrienterNet under all condition variations, even though we train RHO using CV-RHO while evaluating the model using the cross-region test set.
Moreover, the performance drop of RHO is smaller than that of OrienterNet in the cross-region setting.
These results prove the excellent cross-region visual localization ability of our proposed RHO.

\noindent\textbf{Results of Sim2Real Evaluation}.
Apart from generalization, the Sim2Real gap is not negligible since the CV-RHO dataset contains generated adverse images.
We additionally collect a test set with $5$k real-world images under different conditions to study the Sim2Real gap.
All models are trained on CV-RHO while evaluated on both Sim2Real and cross-region test sets.
Since the Sim2Real dataset is collected in the new cities unseen in CV-RHO, we compare the Sim2Real results with the cross-region results instead of the same-region results of the CV-RHO test set for fairness.
All results are listed in Table~\ref{tab:sim2real}.
It can be observed that even though the Sim2Real gap exists, models trained on CV-RHO still achieve high position and orientation recalls, especially on the PR@$5$m and OR@$5$° metrics, proving the value of our proposed CV-RHO dataset and the effectiveness of the RHO model.

\begin{table}[!t]
\centering
\caption{Results of different merge strategies evaluated with the clean data of Washington in our CV-RHO dataset.
Position R and Orientation R stand for Position Recall and Orientation Recall, respectively. All results are in percentages.}
\label{tab:merge-shapes}
\resizebox{\columnwidth}{!}{%
\begin{tabular}{ccccccc}
\toprule[1.5pt]
\multirow{2}{*}{\textbf{FoV}} & \multicolumn{3}{c}{\textbf{Position R @ Xm}~$\uparrow$} & \multicolumn{3}{c}{\textbf{Orientation R @ X°}~$\uparrow$} \\ \cline{2-7} 
     &                         1m    & 3m    & 5m    & 1°    & 3°    & 5°    \\ \midrule
40°  &                         0.29  & 2.82  & 6.39  & 10.40  & 23.63 & 30.68 \\ 
60°  &                         0.34  & 3.04  & 7.34  & 0.90   & 2.55  & 4.07  \\
90°  &                         3.21  & 21.82 & 34.22 & 16.17 & 35.79 & 41.26 \\
\rowcolor{gray!15}120° &       \textbf{14.44} & \textbf{68.14} & \textbf{80.60}  & \textbf{18.61} & \textbf{51.47} & \textbf{69.87} \\
\bottomrule[1.5pt]
\end{tabular}%
}
\end{table}
\begin{table}[h]
\centering
\caption{Analysis of different fusion strategies. Position R and Orientation R stand for Position Recall and Orientation Recall, respectively. All results are in percentages.}
\label{tab:fusion-strategy}
\renewcommand{\arraystretch}{1.0}
\resizebox{\columnwidth}{!}{%
\begin{tabular}{lcccccc}
\toprule[1.5pt]
\textbf{Strategy} &
  \multicolumn{3}{c}{\textbf{Position R @ Xm}~$\uparrow$} &
  \multicolumn{3}{c}{\textbf{Orientation R @ X°}~$\uparrow$} \\ \cline{2-7} 
 & 1m & 3m & 5m & 1° & 3° & 5° \\ \hline

No Fusion &
  14.44 & 68.14 & 80.60 & 18.61 & 51.47 & 69.87 \\

$\text{Prior}_{\text{uv}}$ Fusion &
  11.93 & 59.47 & 77.38 & 30.29 & 73.13 & 85.14 \\

$\text{Prior}_{\theta}$ Fusion &
  13.66 & 68.63 & 81.76 & 32.15 & 78.21 & 88.69 \\

\rowcolor{gray!15}
POF (ours) &
  \textbf{14.53} & \textbf{70.33} & \textbf{83.16} &
  \textbf{37.39} & \textbf{82.96} & \textbf{91.21} \\

\midrule[1.5pt]
\end{tabular}%
}
\end{table}

\subsection{Ablation Study}
\label{sec:ablation}
\noindent \textbf{Analysis of BEV feature shape.}
We analyze various shapes and merge strategies of pinhole BEV features of the SUM module. 
A $120^{\circ}$-FoV pinhole BEV feature is in the shape of an isosceles triangle with a $120^{\circ}$ vertex angle. By rotating the 3 views anticlockwise $0^{\circ}$, $120^{\circ}$, and $240^{\circ}$ respectively, we concatenate them into an equilateral triangle to represent the panoramic BEV feature. 
If a panorama is split into $4$ views, each with a $90^{\circ}$ FoV, the projected pinhole BEV feature is in the shape of an isosceles right triangle.
After rotation, the 4 pinhole BEV feature maps can be merged into a square.
In addition to $120^{\circ}$- and $90^{\circ}$-FoV pinhole BEV features, we also study the split pieces with $60^{\circ}$ and $40^{\circ}$ FoV.
All results are reported in Table~\ref{tab:merge-shapes}.
Results in Table~\ref{tab:merge-shapes} show that an equilateral triangle consisting of three $120^{\circ}$-FoV BEV features outperforms all other merged shapes with a performance gain up to $74$\% in PR@$5$m. This proves that our SUM strategy, splitting a panorama into three $120^{\circ}$-FoV pinhole images, projecting them into BEV features, and then merging them, is effective for modeling a panoramic BEV feature.

\noindent \textbf{Analysis of fusion strategy.} We found that although the merged panoramic BEV feature achieves better results in position recall, the pinhole BEV feature is more helpful in predicting orientation angle.
The reason for these findings is that panorama contains more visual information compared with pinholes, which is beneficial for predicting camera position.
However, the Shannon Entropy remains the same when a panoramic image rotates by an angle.
The case is different when leveraging a pinhole image to predict the orientation angle.
The visual contents before and after the rotation are different in a pinhole image, resulting in different Shannon Entropy.
This leads us to our fusion strategy, which combines the remarkable positioning ability of panoramic BEV features with the precise orientation prediction of pinhole BEV features. 
In Table~\ref{tab:fusion-strategy}, we explored several possible strategies that could be suitable for score volume fusion.
Firstly, we try to extract the precise position prior from the panorama score volume $\mathbf{S}_{pano}$ and add it directly to the pinhole score volume $\mathbf{S}_{1}$.
This is an intuitive approach to fuse them together.
We denote this fusion as $\text{Prior}_{uv}$ Fusion in Table~\ref{tab:fusion-strategy}.
Secondly, we try to extract the accurate orientation prior from $\mathbf{S}_{1}$ and add it to $\mathbf{S}_{pano}$, denoted as $\text{Prior}_{\theta}$ Fusion.
Results in Table~\ref{tab:fusion-strategy} show that taking $\mathbf{S}_{pano}$ rather than $\mathbf{S}_{1}$ as the base score volume to estimate the final pose achieves higher recall both in position and orientation.
Taking $\mathbf{S}_{pano}$ as the base score volume boosts the orientation recall up to $88.69\%$ at the threshold $5$°.
This suggests that the $\mathbf{S}_{pano}$ should be the main score volume.
However, these two fusion methods involve only rectification on one score volume.
We want $\mathbf{S}_{1}$ and $\mathbf{S}_{pano}$ to influence each other mutually. 
Therefore, we design our final fusion strategy to extract the position prior from $\mathbf{S}_{pano}$ first, and use it to rectify $\mathbf{S}_{1}$.
Next, the rectified $\mathbf{S}_{1}$ produces an accurate orientation prior and is used to correct $\mathbf{S}_{pano}$. 
Each fusion stage has a learnable parameter to manage the weight of priors. 
Results in Table~\ref{tab:fusion-strategy} confirm our Position-Orientation Fusion strategy maximizes the strengths of panoramas and pinholes in terms of position and orientation prediction.
Our POF module demonstrates orientation recall improvements of up to $20\%$ compared to using $\mathbf{S}_{pano}$ with no fusion.
Meanwhile, the position prediction preserves its accuracy and yields a moderate performance enhancement.

\section{Conclusion}
In this work, we introduced CV-RHO, the first large-scale benchmark for evaluating the robustness of OSM-based MCVGL across diverse weather, lighting, and sensor conditions. 
With over $2.7$M images from multiple cities and eight realistic variations, CV-RHO enables comprehensive assessment beyond clean settings.
Building on this dataset, we proposed RHO, a robust metric geo-localization model with a two-branch Pin–Pan architecture. The panoramic branch leverages $360${\textdegree} imagery with our SUM module to correct projection distortion, while the pinhole branch provides precise orientation cues. Our POF mechanism effectively fuses the strengths of both branches. Experiments demonstrate that RHO surpasses existing methods and maintains strong performance under challenging conditions, showing solid generalization to adverse real-world scenarios and unseen cities. 
Future work should focus on developing more sophisticated model architectures capable of processing multi-scale datasets and extending the two-branch framework to accommodate diverse input modalities. The evaluation across various datasets should be conducted to test the model’s generalization capability

\clearpage

\section*{Acknowledgments}
This work was supported in part by National Natural Science Foundation of China under Grant No. 62503166 and No. 62473139, in part by the Hunan Provincial Research and Development Project (Grant No. 2025QK3019), in part by the State Key Laboratory of Autonomous Intelligent Unmanned Systems (the opening project number ZZKF2025-2-10), in part by the Deutsche Forschungsgemeinschaft (DFG, German Research Foundation) - SFB 1574 - 471687386, and in part by Helmholtz Association of German Research Centers, in part by the Ministry of Science, Research and the Arts of Baden-W\"urttemberg (MWK) through the Cooperative Graduate School Accessibility through AI-based Assistive Technology (KATE) under Grant BW6-03, and in part by the Helmholtz Association Initiative and Networking Fund on the HAICORE@KIT and HOREKA@KIT partition. This research was partially funded by the Ministry of Education and Science of Bulgaria (support for INSAIT, part of the Bulgarian National Roadmap for Research Infrastructure).

{
    \small
    \bibliographystyle{ieeenat_fullname}
    \bibliography{main}
}

\clearpage
\setcounter{page}{1}
\maketitlesupplementary

\appendix 

\section{More Implementation Details}
A batch size of $36$ was employed when training exclusively with a single \textbf{CV-RHO} variation. In contrast, a larger batch size of $96$ was utilized for training involving all noisy variations or the complete set of data variations. We limit the maximal training epoch to $20$, and the best model checkpoint is selected with early stopping based on validation loss. The evaluation batch size is set to $3$. This value represents the minimal batch size for the RHO model, as the SUM module requires the input of at least three pinhole images for correct operation.

\section{Prompt for Data Generation of CV-RHO} 
\label{sec:prompt}
To achieve a realistic simulation of real-world scenarios under diverse weather conditions without destroying the original buildings and streets' structure in the image, we carefully design prompts and test parameters listed in Table~\ref{tab:prompt2} for generating \textbf{CV-RHO} images. 
All weather variations are generated with seed value $42$ and true cfg scale $1.2$, in inference steps $50$. 
The guidance scale value differs among the $4$ variations.

\begin{figure*}
    \centering
    \includegraphics[width=0.8\textwidth]{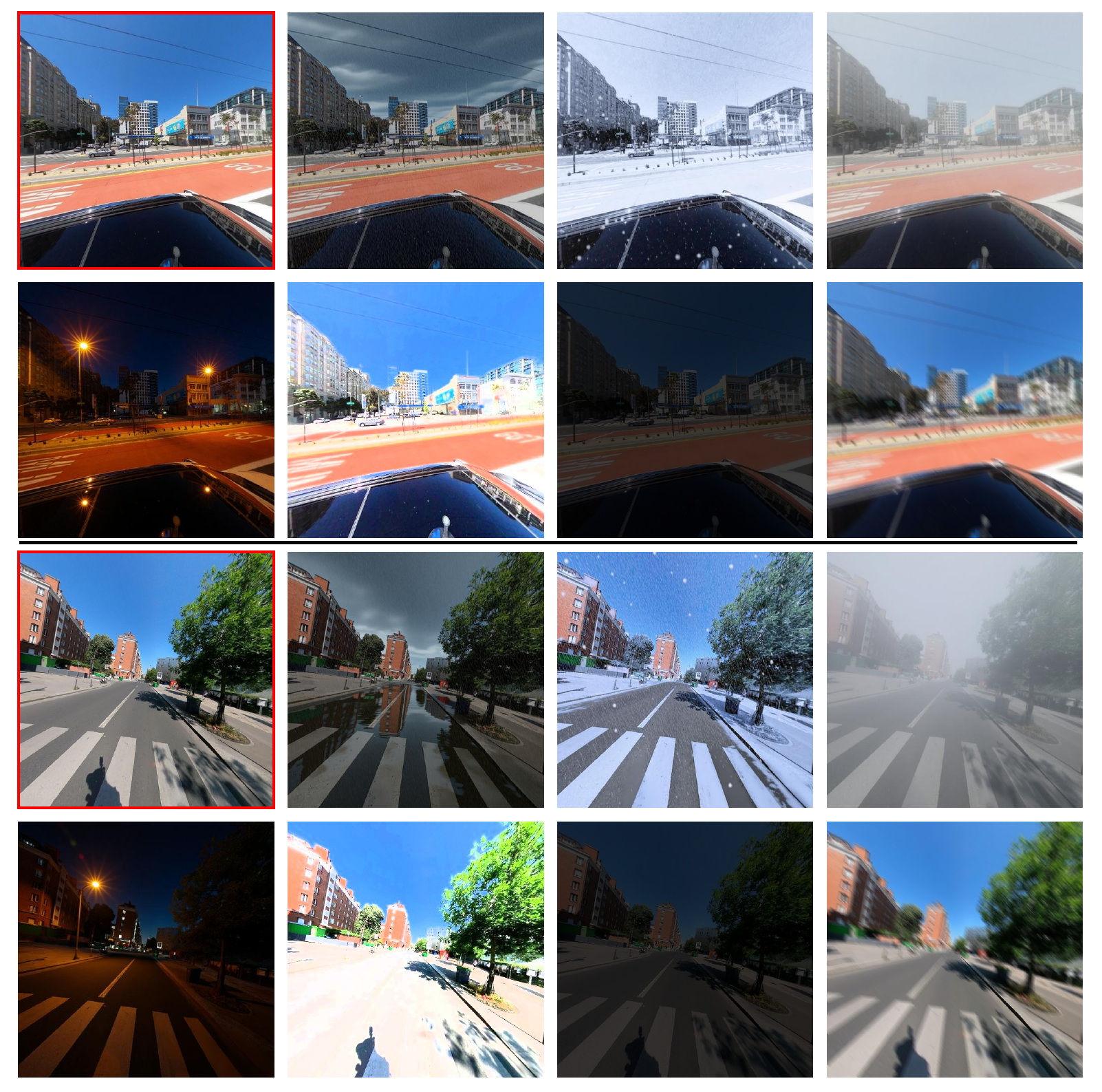}
    \caption{More samples of generated images. Images with red borders are original images.}
    \label{fig:other_data_samples}
\end{figure*}

Utilizing the SOTA and open-source image generation model FLUX.1 Kontext, we edit images to produce photorealistic variations under diverse weather conditions. Fig.~\ref{fig:other_data_samples} presents additional samples of generated images. The generated weather effects are various, rather than being static or uniform, which corresponds to the randomness in the real world.
\begin{table*}[!b] 
\centering
\renewcommand{\arraystretch}{1.2} 
\small 
\caption{Prompts and Parameters configuration for generating images with Flux.1 Kontext.}
\label{tab:prompt2}

\begin{tabularx}{\textwidth}{l X l} 
\toprule[1.5pt]
\textbf{Variation} & \textbf{Prompt} & \textbf{Hyperparameter} \\ 
\midrule

Rain & Change the background into a heavy rainy day, change the sky into gray overcast sky, add visible rain streaks and puddles. & Guidance scale: 7.5 \\
Snow & Change the background into a snowy day, add visible falling snowflakes & Guidance scale: 7.7 \\
Fog & Turn the background into a foggy day, add visible smog on the street & Guidance scale: 7.7 \\
Nighttime & Change the background into evening, illuminate the scene brighter with streetlights. & Guidance scale: 5.0 \\
Over Exposure & - & Brightness factor: 2.5 \\
Under Exposure & - & Brightness factor: 0.25 \\
Motion Blur & - & Kernel size: 10 \\ 

\bottomrule[1.5pt]
\end{tabularx}

\end{table*}

\section{Data Collection of CV-RHO}
\begin{figure*}[!b]
    \centering
    \includegraphics[scale=0.5]{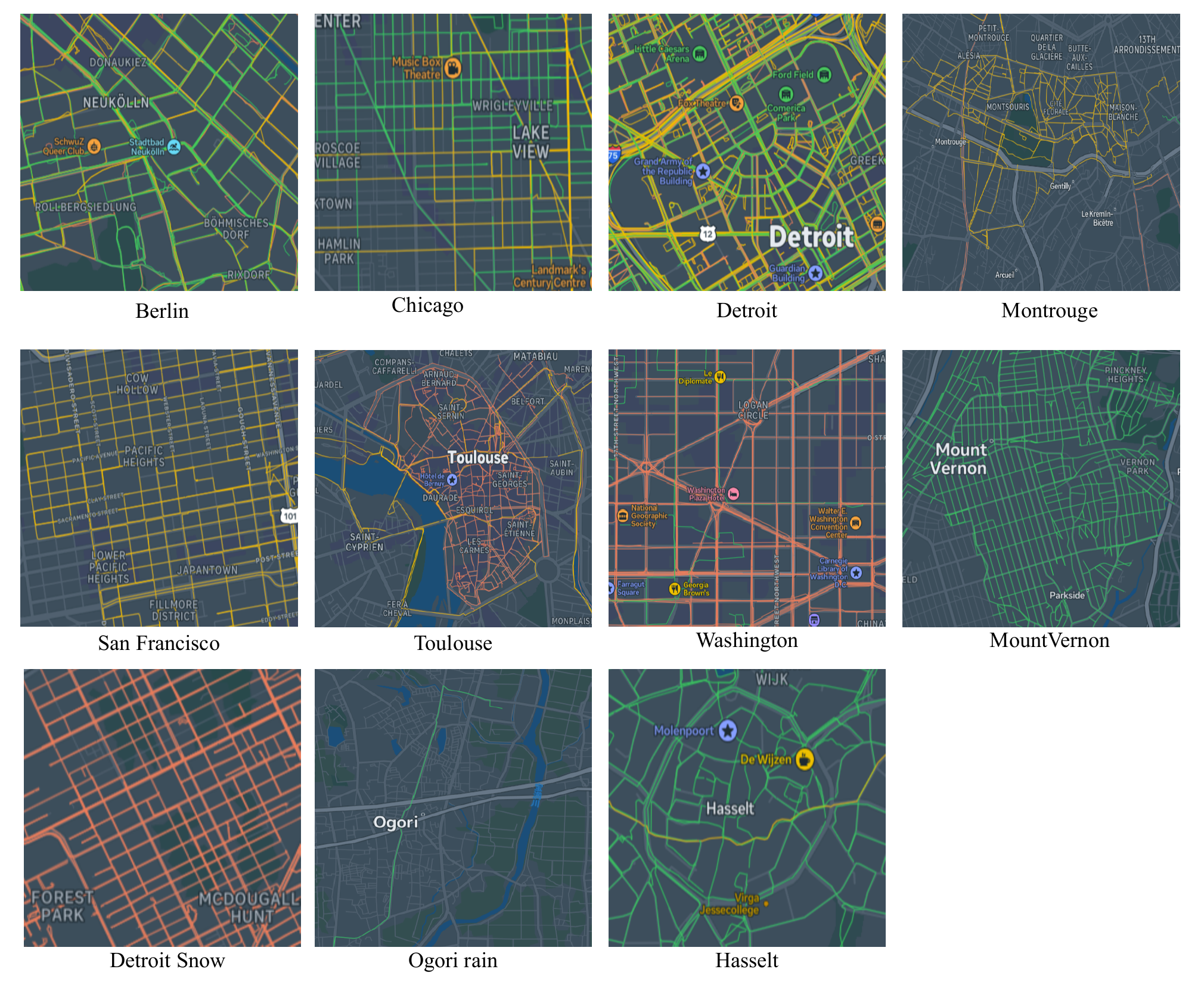}
    \caption{{Selected data points of our CV-RHO dataset and Sim2Real} across $11$ cities. Color from red to green indicates capture date.}
    \label{fig:datapoints}
\end{figure*}

We selected sequences captured after 2020 using panoramic cameras. Fig.~\ref{fig:datapoints} shows OSM covered by the selected sequences. Images of each location were split into disjoint training and validation sets; the ratio between them is approximately $8{:}2$, resulting in \textbf{2.16M} training and \textbf{540K} validation views. Data collected in Mount Vernon is used for cross-region evaluation. In the \textbf{Sim2Real} dataset, the Detroit snow area does not overlap with the area where clean data is collected. Data collected in Hasselt is utilized for sensor noise in \textbf{Sim2Real}.

\section{Detail of Fusion Strategies}
\begin{figure*}[!t]
    \begin{subfigure}[b]{0.42\textwidth}
        \centering
        \includegraphics[width=\textwidth]{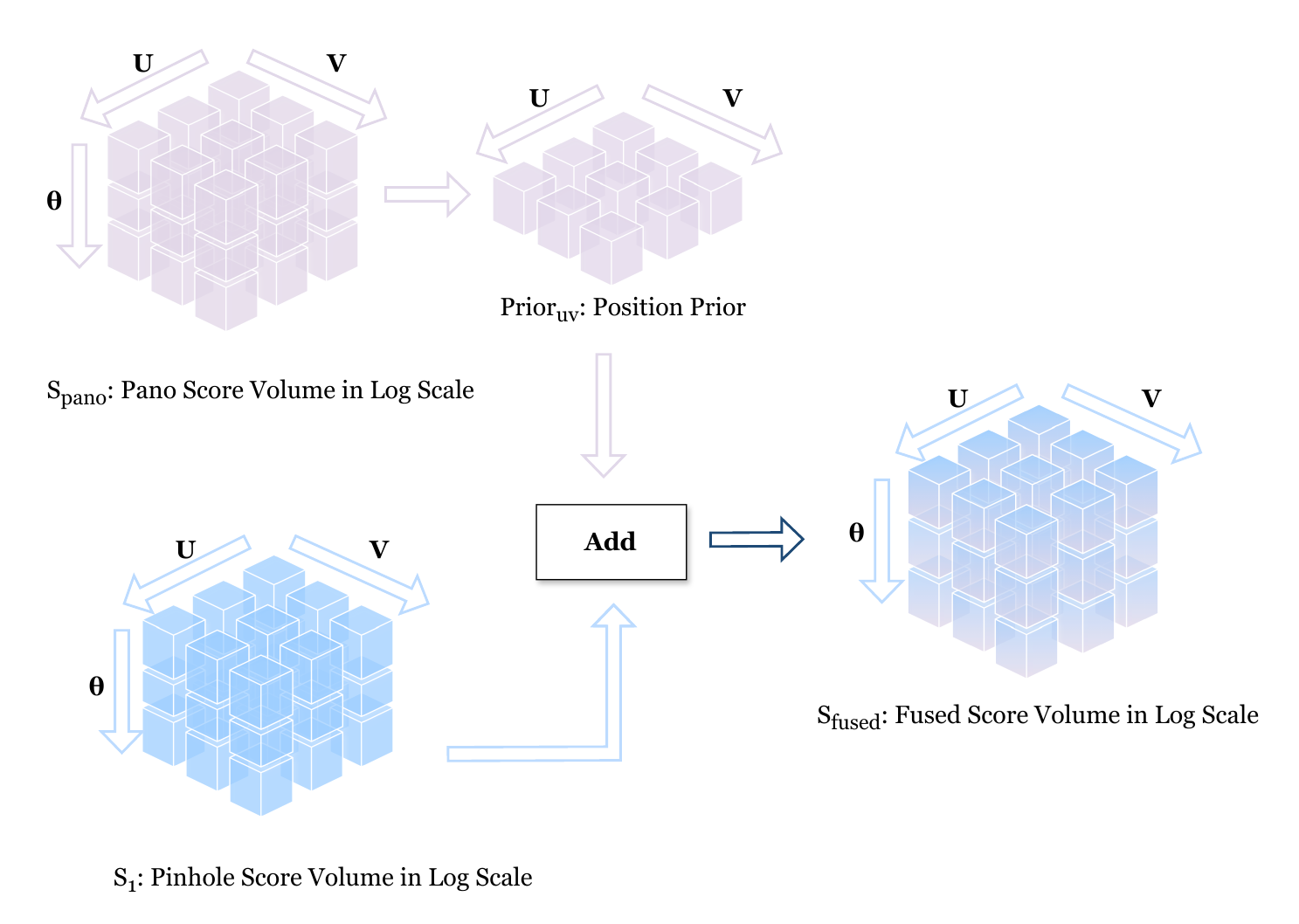}
        \caption{$Prior_{uv}$ fusion}
        \label{fig:position-fusion}
    \end{subfigure}
    \hfill
    \begin{subfigure}[b]{0.48\textwidth}
        \centering
        \includegraphics[width=\textwidth]{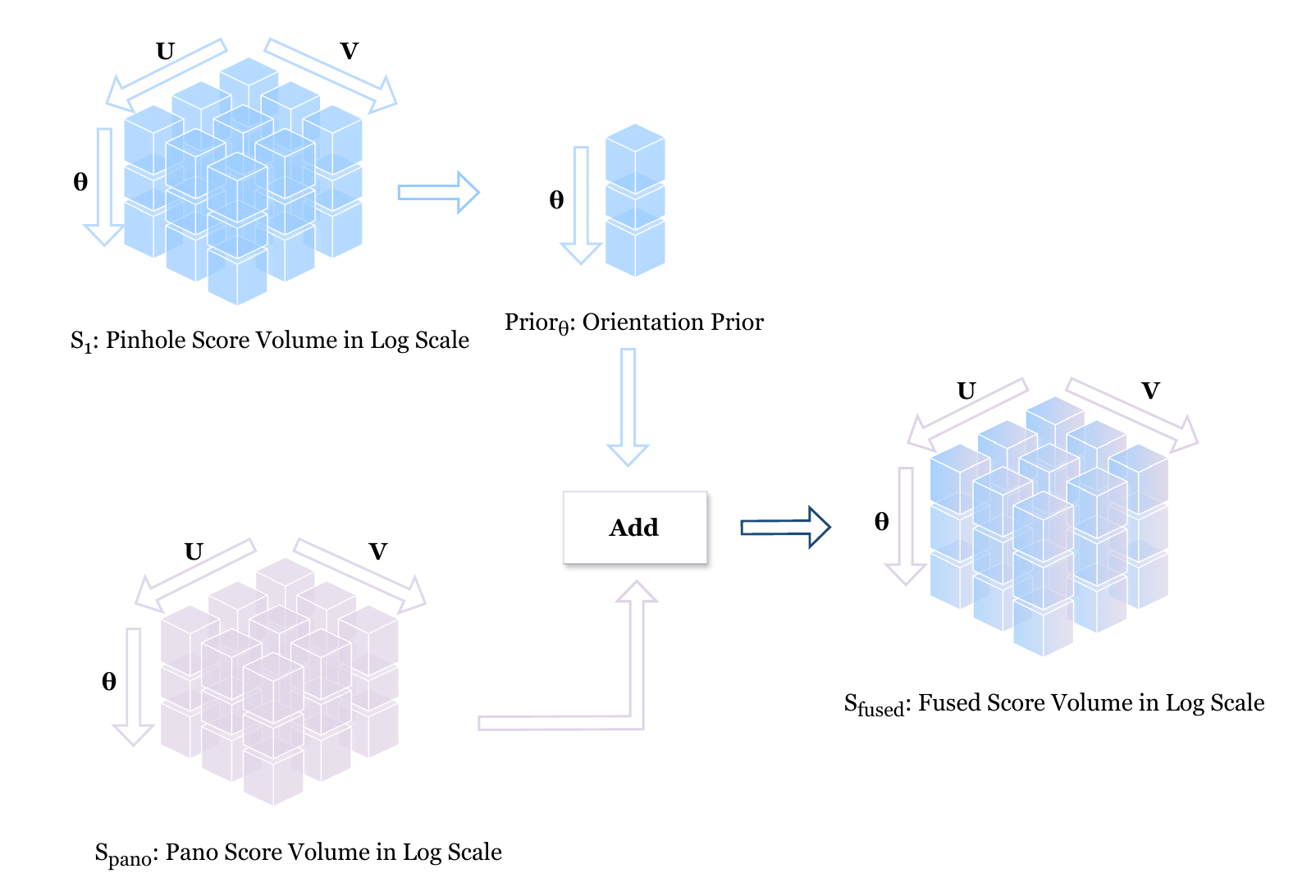}
        \caption{$Prior_{\theta}$ fusion}
        \label{fig:orientation-fusion}
    \end{subfigure}
    \caption{One-way fusion strategy: (a) $Prior_{uv}$ fusion: fuse $S_{1}$ with position prior $Prior_{uv}$. (b) $Prior_{\theta}$ fusion: fuse $S_{pano}$ with orientation prior $Prior_{\theta}$.}
    \label{fig:one-way-fusion}
\end{figure*}

Table~\ref{tab:fusion-strategy} of our main paper shows the performance of different fusion strategies. The \textbf{POF} fusion strategy outperforms $Prior_{uv}$ fusion and $Prior_{\theta}$ fusion, which are components of \textbf{POF} itself. These two fusion strategies are illustrated in Fig.~\ref{fig:one-way-fusion}. 
In $Prior_{uv}$ fusion, the orientation dimension is marginalized out, and the positional probability distribution $Prior_{uv}$ is derived from the panoramic score volume $S_{pano}$. This distribution is then directly added to the score volume $S_{1}$ to refine its position prediction.
In contrast, $Prior_{\theta}$ fusion marginalizes out the positional dimensions to extract an orientation prior from $S_{1}$, which is then incorporated into $S_{pano}$.
If no fusion strategy is employed, we use $S_{pano}$ directly to estimate the final pose.

\section{Limitations and Future Work}

While our model demonstrates remarkable performance on the OSM-based MCVGL task, our proposed SUM module and cross fusion logic are specifically designed for panoramic imagery processing. When applied to pinhole images alone, these mechanisms fail to boost performance compared to pinhole-branch architectures. 
This limitation restricts the model's applicability to scenarios where only pinhole images are available. When training on the complete CV-RHO dataset containing over $2.7M$ images across all variations, our model achieves optimal checkpoints within early steps, despite iterative optimization of learning rates and optimizer configurations. This suggests that the current model architecture may not fully exploit the rich diversity of training samples.

Future work should focus on developing more sophisticated model architectures capable of processing multi-scale datasets and extending the two-branch framework to accommodate diverse input modalities. The evaluation across various datasets should be conducted to test our model's generalization capability.

\end{document}